\documentclass[11pt,a4paper]{article}
\usepackage[hyperref]{naaclhlt2018}
\usepackage{times}
\usepackage{latexsym}

\usepackage{url}

\aclfinalcopy 

\title{Towards NLP with Deep Learning: Convolutional Neural Networks and Recurrent Neural Networks for Offensive Language Identification in Social Media}

\author{Andrei-Bogdan Puiu \\
  Imperial College London \\
  {\tt ap8415@imperial.ac.uk} \\\And
  Andrei-Octavian Brabete \\
  Imperial College London \\
  {\tt ab7515@imperial.ac.uk} \\}

\date{}

\begin{document}
\maketitle
\begin{abstract}
  This short paper presents the design decisions taken and challenges encountered in completing OffensEval 2019 (SemEval 2019 - Task 6) \cite{OLID} \cite{offenseval} , which poses the problem of identifying and categorizing offensive language in tweets. Our proposed solutions explore Deep Learning techniques, Linear Support Vector classification and Random Forests to identify offensive tweets, to classify offenses as targeted or untargeted and eventually to identify the target subject type.
\end{abstract}

\section{Introduction}
The first step we took in tackling this NLP task was to understand the input. This is important to identify and apply the appropriate preprocessing techniques. The next step is to define ways to represent the features based on the language and the diversity of words, such that pre-trained embeddings (e.g., GloVe \cite{glove}, word2vec \cite{w2v}) can be explored or, alternatively, frequency-based or look-up embeddings can be used. We then devise classifier models and validate their performance using the hold-out and cross-validation methods, using the macro-F1 and accuracy metrics. Hyper-parameter optimization is done using grid-search.

\section{Data Preprocessing} 	

The preprocessing step aims to eliminate redundant symbols in tweets and to normalize the text, such that words that convey most meaning are included in the latent representation used as input to the classifiers.

\subsection{Data Cleaning through Regular Expressions}
A number of regular expressions have been compiled to remove emojis, hashtags, user references and non-alphabetic characters such as numbers and ampersands. 

\subsection{Stemming}
Stemming is the process of transforming inflected or derived words to their root form. It is a well-studied problem in machine-based language understanding. The reason for choosing to use such normalization technique is that English language has a high degree of inflection, which can have a visible effect on the feature extraction phase: looking up pre-trained embeddings for non-root words may fail and count-based feature extractors may yield spurious frequencies that incorrectly change the  \textit{weight} of some words.
For our purpose we, are using the Porter Stemmer \cite{porter} from the \texttt{nltk} \cite{nltk} package. This is an alogorithmic approach to transforming words into root forms, but it does not take into account linguistic rules, which may constitute a downside, as sometimes invalid English words are generated.

\subsection{Lemmatization}
Lemattization \cite{nltklemmatization} is the \texttt{nltk} \cite{nltk} provided transform that reduces words to valid roots based on the \textbf{Wordnet} \cite{wordnet} dictionary. We choose to lemmatize only verbs, as this part of speech can provide the most meaning in dynamic contexts found in subtle targeted insults. Lemmatization and stemming, applied exclusively, show significant gains over the case of non-root reductions when used in recurrent models.

\subsection{Tokenization}
The cleaning stage produces the corpus, which is then tokenized \cite{tokenization} by splitting each preprocessed tweet into words. The tokenized corpus is used to create the vocabulary. This is useful to further create the mappings \textbf{word-to-index} and \textbf{index-to-word}, facilitating direct creation of the look-up embedding used as input to the Deep Learning models. We use the tokenized corpus to determine the maximum preprocessed tweet length and to pad to length (with zeros) the array of indices constituting the training, validation and test sets.

\section{Type of Input Data} 	

\subsection{Term-Frequency Inverse Document Frequency (TF-IDF)}
One of our initial assumptions was that word frequency can be indicative of the corresponding class. For example, samples labelled as offensive are more likely to have specific (bad) words with a higher local importance. The same is valid for samples labelled as non-offensive. The intuition can be extended beyond Task A. Feature engineering reduces to finding signature words for tweets through TF-IDF \cite{idf} \cite{tfidf}. We use \textbf{term frequency} to measure the importance of a word in a tweet. A signature word \cite{tfidfarticle} for a tweet should not appear too often in other tweets, thus its \textbf{inverse document frequency} must be high. The formula for calculating smooth TF-IDF for a term/word $t$ in a tweet is given by:

\[
    TF-IDF(C, t) = TF(t) \cdot IDF(C, t) \]\[
    IDF(C, t)   = \log{ \frac{n}{DF(C, t) + 1}}
\]

\hspace*{-0.4cm}where C is the corpus, n is the total number of tweets in the corpus, t is the current term, TF(t) is the frequency of term t in the tweet, DF(C, t) is the frequency of term t in the corpus.

\subsection{Look-up Embedding Layer}
An alternative feature extraction method is to use the index of each word in the vocabulary to retrieve a unique trained embedding of that word from the trainable embedding tensor which acts as a look-up table. We use the \texttt{word2index} mapping constructed on the vocabulary array to map each tokenized tweet to an array of indices that represent keys for retrieving the learnt embedding.

\section{Classifiers} 	

Our motivation for the choice of classifiers is presented next. The Convolutional Neural Network and the Recurrent Neural Networks (LSTM and GRU-based) are implemented in PyTorch \cite{torch} and the Linear Support Vector Classifier, Logistic Regression and Random Forest use the Scikit Learn library. The descriptions of architectures closely follow the Neural Network API for layers and activations, as provided by the PyTorch framework.

\subsection{Convolutional Neural Network}
CNNs \cite{cnn} are the go-to method in Deep Learning for classification when fast training is required. Implementation of convolutions is optimized on modern CPUs and GPUs by vectorized instructions, in contrast to more effective but slow linear computations required by RNNs based on LSTMs and GRUs. Our CNN model is composed of an embedding layer, a convolutional layer activated by ReLU, a max pooling layer, a dropout layer and a linear layer. It is worth explaining the mechanics behind the convolutional layer when the input is a set of word embeddings representing one tweet from the corpus. This can be best associated with the n-gram model, where a sliding window of size n is run across the whole sentence. Given the word embeddings, each n embedding vectors are convolved with the filters, giving 1-dimensional activation maps next pooled by the max-pooling layer. This mechanics intrinsically takes into consideration context, so simpler input preprocessing suffices. The dropout layer plays an important role, as such model can exhibit neuron co-adaptation which can degenerate through high activations of irrelevant word associations given by the n-gram.

\subsection{Recurrent Neural Network - Long Short-Term Memory (LSTM)}

Starting from the fact that language is sequential and maintains this property even after preprocessing, we decided to explore Recurrent Neural Networks for the given tasks. Given that plain RNNs suffer from the vanishing and exploding gradient problems, which would make experimentation with Deep RNNs difficult, we opted for the Long Short-Term Memory unit \cite{lstm} to learn the underlying structure of the sequential inputs (words). For example, for a phrase such as "NLP is fun", the sequential input is represented by the words "NLP", "is" and "fun".

The mechanism on which LSTM cells are based involves three gates that determine the cell state. The state comprises the output used by superior layers and the hidden state passed to other LSTM units. Information removed or added to the cell is controlled by three gates: the input gate controls how much to write to the cell, the forget gate controls how much to erase or keep in the cell and the output gate how much to output from the cell. The weights associated to these gates are updated through \textit{back-propagation through time}.

In our design, we decided to adapt the architecture used for the Convolutional Neural Network case starting with an embedding layer of size 100 and replacing the convolutional layer with one LSTM unit with outputs of dimension 32. The next layer is a linear layer activated with ReLU. Predictions are obtained using a softmax layer which outputs probabilities, which makes it suitable for probabilistic interpretation in classification tasks.

\subsection{Recurrent Neural Network - Gated Recurrent Unit (GRU)}
A Gated Recurrent Unit is similar to an LSTM and has proven to be as effective as an LSTM, sometimes surpassing its performance for certain tasks. GRUs \cite{gru} have become solutions of choice because of computational efficiency which makes training faster. The GRU is composed of and update gate and a reset gate. The interesting part of GRUs is the \textit{decision} of update gates on the \textit{quantity} of information to be passed on to new gates or to other layers. GRUs can withhold information from further time steps in the past, without completely forgetting information as LSTMs do \cite{gruarticle}. 

\subsection{Scikit Learn Models}

We used the \textit{Scikit Learn} library \cite{scikit-learn} to see how well classifiers that are not convolutional or recurrent neural networks perform on this task. We evaluated all the main classifiers offered by the library, and selected the following three classification algorithms to be part of our submission, based on their performance on the task:

\textbf{Linear SVM Classifier} The library provides an implementation of support vector classification using a linear kernel. The underlying implementation is a quadratic programming problem that solves the primal optimization problem, which has as many variables as the number of features in the data set.

\textbf{Logistic Regression} Our next classification model is a based on the logistic regression model, as implemented in the Scikit Learn library. The two-class and multi-class training are provided through the API \cite{sklearn-api}, which uses the \textit{liblinear} \cite{liblinear} solver and multiple optimization techniques.

\textbf{Random Forest} This ensemble learning method is based on a number of distinct trees built from different random samples from the dataset, which prevents overfitting. The basic principle of training decision trees is to favor the features which have the highest information gain and to use those to best distinguish between two classes. Unlike for the neural network approach, here we did not have an embedding layer to transform our sentences to vector form. Instead, we used \textbf{TF-IDF} to obtain our features. The random forest approach groups together a number of trees on which predictions are made. The mode of those predictions is chosen as the final prediction.

\section{Performance}

\subsection{Validation Experiments}

\begin{table}[ht!]
\centering
\begin{tabular}{llll}
  Task & CNN & LSTM  & GRU\\
  \hline
  A & 0.5573 & 0.6137 & 0.5819 \\
  B & 0.9114 & 0.5509 & 0.5300 \\
  C & 0.4724 & 0.4476 & 0.4378
\end{tabular}
\caption{Cross-validation average macro F1 measure on top performing models}
\end{table}

We initially started with the hold-out method to get an intuitive understanding of the performance of the models. However, this method does not fully help measure the performance, as only part of the data-set is used. Therefore, we switched to using K-fold cross validation (for $K = 5$), as it is more appropriate for small datasets, as is the case of the tweets dataset, such that all examples get to take part in training/validation. In Table 1, we give the macro F1 score for the best-performing CNN, GRU and LSTM architectures, obtained by averaging over all cross-validation runs.

In supervised learning all training examples
have a target label. It is good practice to have a validation set as well for tuning the hyperparameters,
so in the following explanation we will include this as well. The hold-out method consists of splitting the data set into: training data and validation data. The model is trained on the training data, the set of best hyperparameters chosen on the validation data using the F1 measure. The advantage of
this method is that there is a single pass through the data set, so it is computationally efficient. The disadvantage is that the performance
measures may have high variance, depending on how the data set splitting is done, so the method gives less reliable
statistical measures. K-fold cross-validation works as follows: the data-set is split into k equal folds - k - 1 training folds and
1 validation fold. The model is trained on the k - 1 folds and the validation F1 measure is computed on the validation fold. The process
is repeated until each of the k folds plays the role of the validation fold. Thus, the process is completed after k training iterations. After each
iteration, the parameters of the model are reset and the training restarts on the next k - 1 folds. The performance measures are 
computed by averaging each metric for all k iterations. The advantage of this method is that it provides a more reliable 
statistical measure of the performance of the model, with less variance. The disadvantage is the computational cost to train for all splits.

\subsection{Hyperparameter optimization}
We used grid search to explore the hyper-parameter space of each of our models. For each type of model we implemented, we defined a method \textit{optimize}, which searched a model-specific parameter grid, trained models with all the parameters and produced the best performing model. \\ 
For each of our neural network categories, the way we evaluated model performance was by doing 5-fold cross-validation on the training set, with the \textit{score} of each model being evaluated as the average macro F1 score on the \textit{k} folds; the model with the best score was then selected to be the model of that type. \\
For our other models, we used the \textit{Scikit Learn} class \textbf{GridSearchCV}, which took care of the optimization and found the best model.

\subsection{Submissions Performance}
The F1 measures on the test set for the top three performing models are presented in Table 2. We show that our models are able to achieve a macro F1 measure above the baseline imposed by the competition organizers: 0.4189 for Task A, 0.4702 for Task B and 0.213 for Task C.
The non-neural models tended to perform well on the training set, but generalized worse than the neural models on the test set.

\subsection{Discussion of Results}
We observe that the convolutional neural network provides the best results on the CodaLab test set, outperforming the GRU- and LSTM-based networks by a decent margin. Moreover, these results are consistent with those we obtained in our validation experiments, where the CNN was also found to be the best model. \\
The models as a whole do not obtain outstanding results on the test set, due to the lack of data in the training set, which does not allow them to learn to distinguish between classes at a high level of confidence. The results after hyper-parameter tuning yield best performance for the CNN model. \\

\begin{table}[ht!]
\centering
\begin{tabular}{lllll}
  Task & CNN & LSTM & GRU & Best\\
  \hline
  A & 0.5558 & 0.4968 & 0.4890 & 0.6835\\
  B & 0.5754 & 0.5502 & 0.5168 & 0.6296 \\
  C & 0.3510 & 0.3312 & 0.2842 & 0.4931 \\
\end{tabular}
\caption{Codalab Competition Submission Macro F1 measure Top Performing Models}
\end{table}

\subsection{Challenges in Classifier Design for NLP}
Designing a classifier for Natural Language Processing tasks involves deep understanding of the data-set. This is important to guide input preprocessing and feature extraction, steps of considerable importance, as classifiers are prone to achieving highest information gain from non-spurious data. Most often, heuristics are sufficient to decide what should be eliminated from the input in alignment with the classification task goal.

One notable challenge is to decide on the type of embedding used as input to the classifier. This involves deciding whether the contexts in which specific words are found may increase the likelihood of placement in one of the two classes. In such a case, words should be represented by their context which can be achieved by training a neural network in a skip-gram model context and pulling the trained weights which act as the embedding. This is a predictive model known as \textbf{word2vec}. In contrast, count-based models learn the word embeddings by doing dimensionality reduction on the co-occurence counts matrix, which is usually of dimension quadratic in the vocabulary size. Features constructed when the semantic context is not taken into account (e.g, TF-IDF), may have downsides especially when the task can benefit from context-aware embeddings. Thus, with the large number of choices of feature representation, the practitioner needs to carefully ponder and eventually select the one that is most suitable for the task. For example, after analyzing the tweets we decided that offenses are predominantly characterized by specific words and thus selected context-unaware approaches.

After exploring multiple options in what we consider the most important steps to a successful natural language classification, input preprocessing and feature engineering, the remaining task is to devise the classifier. Our approach was to pick models that were expected to leverage certain characteristics of the data, to optimize them independently and to compare test set results. We started from simple models and made narrow assumptions such as linear separability of the data, thus devised a Linear Support Vector Classifier. Such a simple model achieves test set performance close to the baseline, but its learning capacity is insufficient. Next, a simple logistic regression model achieves test metrics above the baseline, which was unexpected in the incipient phases of solving the task. Given no previous experience in NLP tasks, the incremental approach helped us better understand the flaws of some models and to devise models with much higher learning capacity such as CNNs and RNNs.

One other challenge was to train models on imbalanced data sets such as the one for task B, where there are significantly fewer examples of untargeted offenses and to ensure that test data belonging to this class are correctly classified. We achieved this with data augmentation, by creating new UNT examples through random tweet construction from words randomly picked from already present examples.

\section{Conclusion}
The project proved to be challenging, but it highlighted some key, fundamental concepts from the Natural Language Processing universe. Starting from simple models and gaining more understanding of the dataset and how it can be best represented, we eventually achieved notable performance improvements through refining the design choices for the classifiers. \\

\bibliography{semeval2018}
\bibliographystyle{acl_natbib}

\end{document}